# Real-time Attention Span Tracking in Online Education

Rahul RK
2017it0638@svce.ac.in

Shanthakumar S
2017it0702@svce.ac.in

Vykunth P
2017it0744@svce.ac.in

Sairamnath K
2017it0753@svce.ac.in

*Department of Information Technology, Sri Venkateswara College of Engineering, Sriperumbudur (Autonomous, Affiliated to Anna University)*

*Abstract*—Over the last decade, e-learning has revolutionized how students learn by providing them access to quality education whenever and wherever they want. However, students often get distracted because of various reasons, which affect the learning capacity to a great extent. Many researchers have been trying to improve the quality of online education, but we need a holistic approach to address this issue. This paper intends to provide a mechanism that uses the camera feed and microphone input to monitor the real-time attention level of students during online classes. We explore various image processing techniques and machine learning algorithms throughout this study. We propose a system that uses five distinct non-verbal features to calculate the attention score of the student during computer-based tasks and generate real-time feedback for both students and the organization. We can use the generated feedback as a heuristic value to analyze the overall performance of students as well as the teaching standards of the lecturers.

## I. INTRODUCTION

The demand and need for online education are increasing rapidly. Almost all the schools and colleges throughout the world have shifted to the online mode of lectures and exams due to the recent coronavirus outbreak, and this trend will most likely continue in the upcoming years. The increasing demand for online education opens the gate to automation in the field. One major issue in the online mode of lectures is that students tend to lose their concentration after a certain period and there is no automated mechanism to monitor their activities during the classes. Some students tend to just start a lecture online and move away from the place, or might even use a proxy to write online tests for them. This situation also takes place in online course platforms such as EdX and Coursera where the student tries to skip lectures just for the sake of completion and certification. The loss in concentration not only affects the student's knowledge level but also hurts the society by producing low-skilled laborers. We propose a solution in our paper to address this issue.

The paper is structured as follows: Section 2 reviews the literature, Section 3 describes the proposed methodology and its working, Section 4 evaluates the performance, and Section 6 concludes the paper and talks about ideas for future works.

## II. LITERATURE REVIEW

The purpose of attention span detection during online classes is to gather data and analyze the state of the student, to evaluate his performance based on concentration level, instead of just academic scores. According to [1], the average blink rate of a person is between 8 to 21 blinks per minute, but when the person is deeply focused on a specific visual task, the rate of blinking has significantly reduced to an average of 4.5 blinks per minute. Likewise, the blink rate escalated to over 32.5 blinks per minute when the individual's concentration level is low. The study [2] explores how the emotional state of students varies during the learning process and how emotional feedback can improve learning experiences. Emotional states such as happiness, joy, surprise, and neutral denote a positive constructive learning experience, whereas, emotions like sadness, fear, anger, and disgust represent a negative experience. The study [3] discusses how mind-wandering can have negative effects on performance and how eye gaze data, collected using a dedicated eye tracker, can automatically detect loss of attention during computer-based tasks. The eye behavior patterns were observed and it is found that the patterns varied distinctly during mind wandering.

Article [4] shows how the loss of attention significantly affects the learning efficiency of students. The review paper [5] suggests that drowsiness, caused by sleep, restlessness, and mental pressure, is one of the major factors that lead to loss of attention. Various state-of-the-art drowsiness detection techniques were compared, and it is found that the Haar classifier and Support Vector Machine (SVM) gives better accuracy in real-time scenarios. The study [6] establishes evidence that the attention span of students is contingent upon the environmental noise conditions, and the results suggest that noise levels greater than 75dB have a serious impact on the accuracy of the students. The paper [7] proposes an automated facial recognition model using Convolutional Neural Network (CNN) and Principle Component Analysis (PCA). The study [8] proposes a two-layer CNN to learn the high-level sparse and selective facial feature maps. Sparse Representation Classifier (SRC) improves the performance by using a sparsely selected feature extractor. The study [9] examines the features of body posture and head pose to predict the user's attention level by identifying patterns of behavior associated with attention.

From the literature review, we have identified that the five parameters - blink rate, facial expression, eye gaze, background noise, and body posture make a good feature set to assess the attention level of the students.

## III. PROPOSED METHODOLOGY

This study makes use of five parameters to calculate the attention-span level of the student attending the online class. Facial recognition is used to validate the student's attendance. The attention span score is calculated using blink rate, facial expression, eye gaze, background noise, and body posture and is updated continuously for a window length of 5 seconds. Instead of sequential execution, all the models required to calculate the attention span are executed in parallel once the online lecture starts. This is achieved using multithreading all the functions, which plays a major role in reducing the time consumption of each model as well as the whole system. For every 5 seconds, the model will generate the attention span score and provide real-time feedback to the students in the form of live graphs which are plotted for each parameter as well as the calculated attention span score. The following sections will explain in detail about each of the models used in this study and their significance in calculating the attention span. The overall architecture of the proposed system is shown in (Fig. 1) and the working of each module is represented as flowcharts in (Fig. 2).

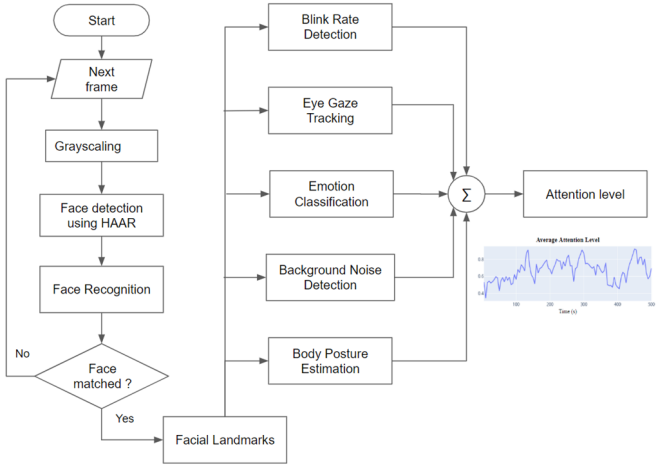

Fig. 1. The architecture of the proposed system.

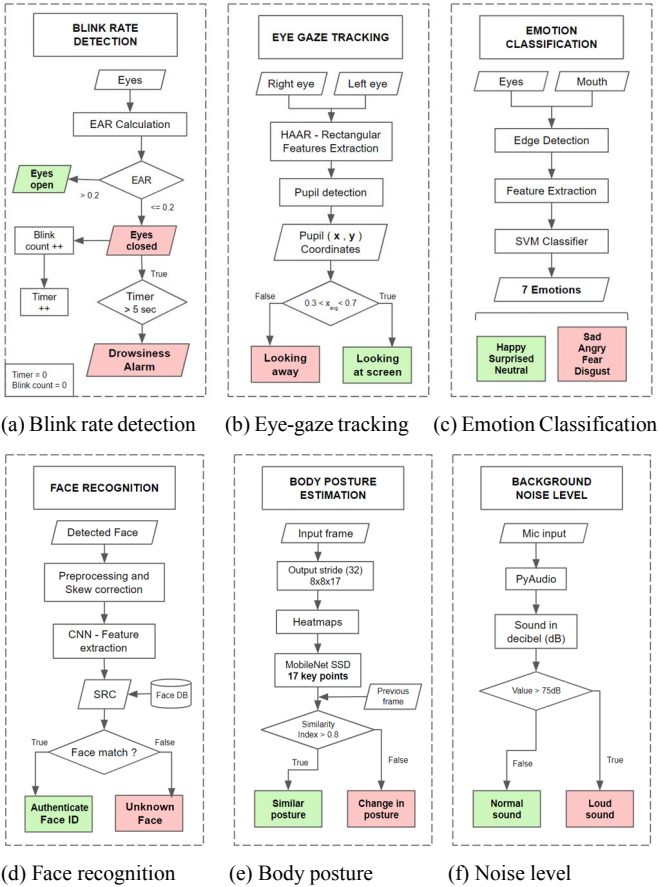

(a) Blink rate detection  (b) Eye-gaze tracking  (c) Emotion Classification

(d) Face recognition  (e) Body posture  (f) Noise level

Fig. 2. Flowcharts of each module

## A. Facial Landmark Detection

Face detection is implemented using the Viola-Jones algorithm [13], which uses a windowing mechanism to scan images for identifying features of human faces. The paper [8] provides an efficient real-time approach to extract 68 key points from the detected face image using OpenCV's dlib library as shown in (Fig. 3). We use rectangular regions to extract Haar features from the image. The landmarks are classified into five categories of facial features: eyebrows, eyes, nose, mouth, and jaw, which are denoted sequentially using the key points. We will be using these individual landmark features as inputs to further modules. We can improve the accuracy of the face detection module by computing more number of key points. However, this increases the processing time.

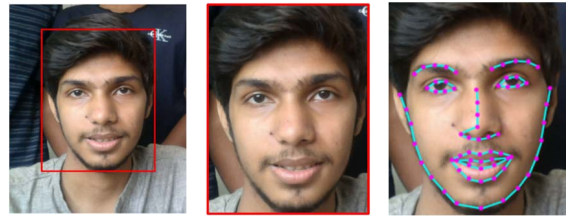

Fig. 3. Face detection and Facial Landmarks.

## B. Blink Rate Detection

Blink rate is one of the important factors to determine the state of mind - whether the student is actively listening or drowsy during the class. In this module, we crop the regions containing the eye pairs and divide each eye into two halves. We calculate the Eye Aspect Ratio (EAR) using Euclidean distances (Fig. 4a) for every frame as per Formula (1) to identify whether the eyes are open or closed. We also have a countdown timer, which is activated once a blink is detected, to keep track of the number of seconds the eyes are closed. It can be concluded that the user is feeling drowsy (loss of attention) if the eyes are found to be closed for more than two seconds [5] and an alert will be given both visually as shown in (Fig. 4c) and warning alarm sounds. We calculate the number of blinks continuously on an interval of 5 seconds to determine the average blink rate of the user. The EAR threshold value is set as 0.2 based on test experiments.

We tested the blink detection module on 306 images comprising of 156 closed eyes and 150 open eyes and classified the blinks with an accuracy of 91.02% and open eyes with an accuracy of 92.66%.

$$EAR = \frac{|p2-p6|+|p3-p5|}{2|p1-p4|} \quad (1)$$

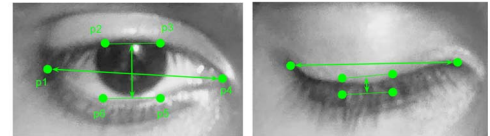

(a) Eye key points  *(p1, p2, p3, p4, p5, p6)*

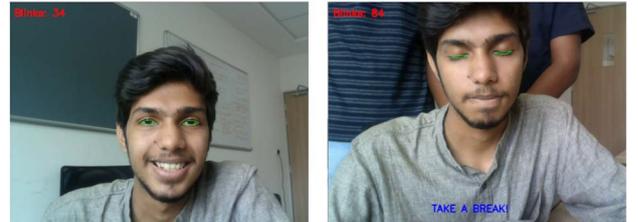

(b) Opened eye  (c) Closed eye and Drowsiness alarm

Fig. 4. Blink rate and drowsiness detection

## C. Eye-gaze Tracking

The eye-gaze of a student is tracked to determine where he is looking at and is often closely associated with the distraction level of the student. As suggested in [10], we analyze the extracted eye region coordinates for rectangular features to identify eye regions containing the pupil. The pupil coordinates (x, y) of each eye (Fig. 5a) are calculated and mapped to determine the eye gaze direction. Based on the resolution of the screen, we have established two possible classes of eye gaze: looking at the screen (Fig. 5c) and looking away (Fig. 5b) (right or left). We collected 150 images with two different classes: looking straight and looking away. Our model was able to classify the eye gaze correctly for 113 images with an accuracy of 75.33%.

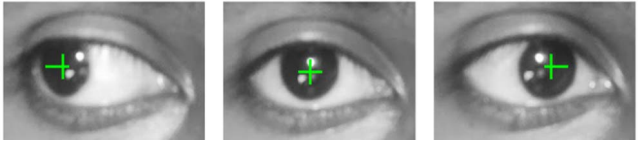

(a) Pupil coordinates – *Left, Center, and Right*.

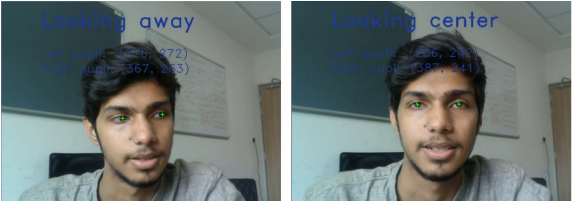

(b) Looking away       (c) Looking center

Fig. 5. Eye-gaze tracking

## D. Emotion Classification

The emotion of the person attending the online class plays a major role in his attention level. This study uses facial features such as eyes, nose, and mouth, extracted using Haar-cascade classifier and facial landmark detector. Support Vector Machine (SVM) algorithm is used to classify the emotion of the students into seven different classes - angry, disgust, fear, happy, sad, surprise, and neutral. A score is given for each emotion depending on its effect on the attention level of the user. The traditional method for emotion classification pre-processing uses only the cropped eye. However, [11] proposed an alternative solution to include mouth features for better accuracy. As this method only uses Haar cascades to classify the emotion, the processing speed was much faster when compared to the Sobel edge eye detection. The model was validated using the JAFFE dataset which contains 213 images and 7 emotions posed by 10 different Japanese women. The training set consisted of 42 and 7 classes of emotion. The test set consisted of 70 images. We obtained an average accuracy of 82.55% on our test set (Fig. 6).

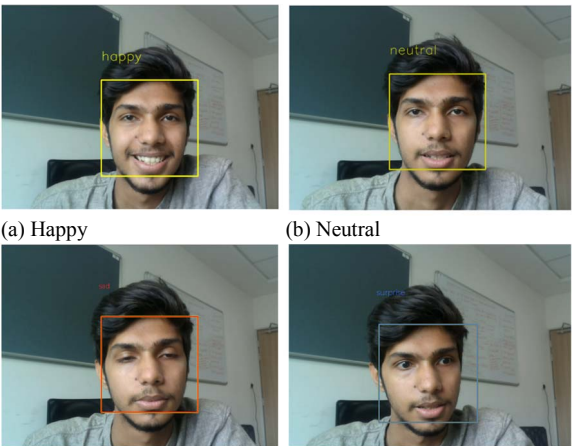

(a) Happy       (b) Neutral

(c) Sad       (d) Surprise

Fig. 6. Emotion Classification *(Happy, Neutral, Sad, Surprise)*

## E. Face Recognition

A robust facial recognition system is essential to authenticate the student based on biometrics to avoid student proxies and to automate the attendance management process using the webcam feed during classes. In this module, we modified the architecture of [8] and implemented a 3 layer Convolutional Neural Network (CNN) consisting of three convolutional layers with max-pooling, a fully connected layer with dropout, and a Sparse Representation Classifier (SRC) output layer. The dropout layer helps in reducing the computational cost. We created a dataset consisting of 500 images with four different faces. We split the dataset into 300 images for training and 200 images for validation. After 15 epochs of training, we attained a training accuracy of 94.8% and a validation accuracy of 90% (Fig. 7).

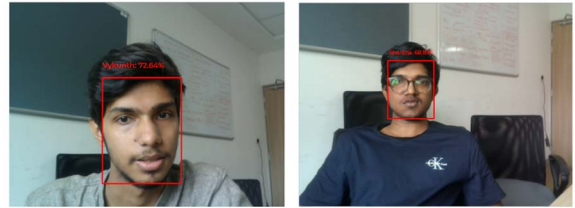

Fig. 7. Face Recognition

## F. Body Posture Estimation

Many researchers have used CNN or R-CNN to estimate the pose with high accuracy. However, the main goal of our research is to calculate the attention level of the student in real-time without compromising on processing time. Hence, we make use of the TensorFlow pose-estimator (PoseNet), based on Mobilenet SSD to estimate the posture of the student. This model uses the heat map to estimate the pose difference from one frame to the previous frame. PoseNet can identify 17 key points including face, shoulders, elbows, wrists, hip, knees, and ankle. We assign a pixel similarity score by comparing the change of head pose and body posture between consecutive frames to predict whether the student is restless or focused during the online lecture.

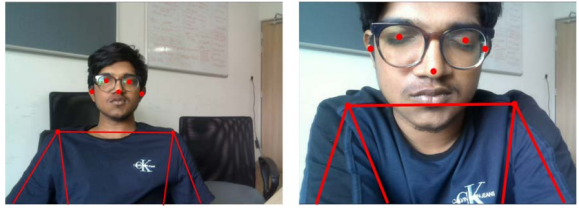

Fig. 8. Body Posture Estimation

## G. Background Noise Detection

We use the python package PyAudio to detect the input audio from the device's microphone. The background noise during the class might affect the concentration level of the student. The average sound level in a school is 50 dB and 75 dB is set as the threshold for loud noise [6]. Anything above 75 dB is considered as a noisy environment and the scores will be inversely proportional to the background noise. The model will monitor the background noise continuously and we calculate the average noise level every 5 seconds.

## H. Overall Attention level Detection

All the scores from the above parameter scores (blink rate detection, eye gaze tracking, emotion classification, body posture estimation, and background noise detection) are normalized to calculate the attention score as per Formula (2). We plot live graphs as shown in (Fig. 9) with the predicted attention level of the student along with the scores for each parameter updated in real-time. We do not use face recognition in the scoring method because it does not contribute to determining the attention level of the student; rather we use it for biometric authentication and automated attendance of the students.

$$Att = \frac{\sum score(i)}{n} * 100 \qquad (2)$$

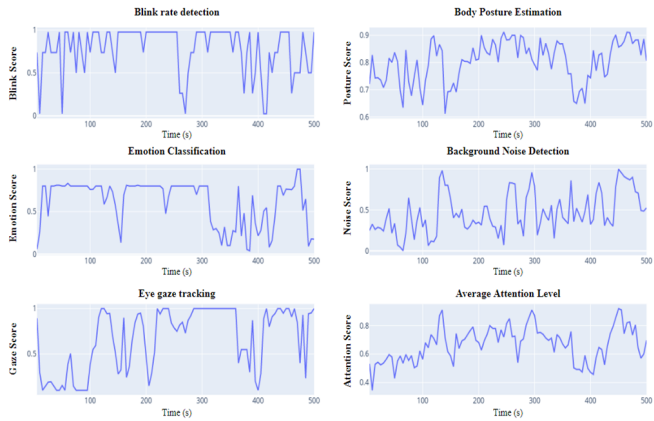

Fig. 9. Live graph plotting with real-time attention level

## IV. PERFORMANCE EVALUATION

The system's performance was analyzed by using a dataset of 15 undergraduate students consisting of nine males and six females. The students were asked to attend online lectures on different topics each for 500 seconds and three human observers were asked to provide an observed attention score based on the recorded web camera video, which is used as the ground truth-value. We compared the predicted attention scores with the observed scores to evaluate the overall performance of our system. (Fig. 10) shows the comparison between predicted scores and observed scores and Table (1) shows the system's performance metrics.

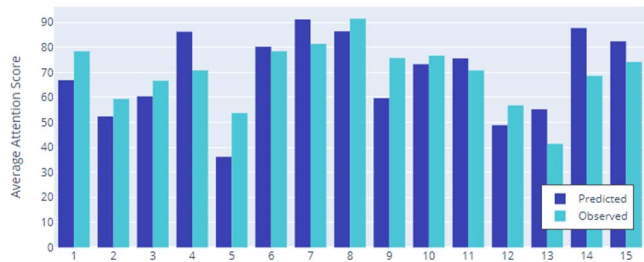

Fig. 10. Predicted Attention Score vs Observed Attention Score Comparison.

TABLE I. PERFORMANCE METRICS

| *Metric* | RMSE | MAE | R2 | MAPE |
|---|---|---|---|---|
| *Value* | 11.152 | 9.837 | 0.154 | 15.248 |

Our system was able to perform quite well given the limited data used for training the models. We obtained the overall accuracy of our attention-tracking model by taking the average of the accuracies of each module. Compiling OpenCV's DNN module and Caffe with CUDA support improved the performance and significantly reduced the inference time of our models as shown in Table (2). We achieved an overall accuracy of 84.6233%.

TABLE II. SYSTEM PERFORMANCE

| *Module* | *Accuracy* | *Inference time* |
|---|---|---|
| Facial Landmarks | 89.67 % | 0.033 ms |
| Blink rate detection | 91.02 % | 0.026 ms |
| Eye gaze tracking | 75.33 % | 0.032 ms |
| Emotion classification | 82.55 % | 0.057 ms |
| Facial recognition | 90.11 % | 0.052 ms |
| Body posture | 79.06 % | 0.048 ms |
| **Overall System** | **84.6233 %** | **0.258 ms** |

## V. CONCLUSION AND FUTURE WORKS

In this paper, we have implemented a system to tackle the issues involved in online education using five parameters. We used the face recognition model to verify the student attending the online class. We used the other five parameters - blink rate, eye gaze, emotion, posture, and noise level to calculate the attention level of the student throughout the lecture. Since this involves real-time processing, we have implemented and used lightweight models to reduce the processing time. We visualize the scores in the form of a live graph and generate automated reports. The feedback generated can be used for:

1) *Evaluating student performance*
2) *Improving teaching standards*
3) *Preventing malpractice during online examinations*

As a part of future works, we can improve our system's performance further by training our models using more data. Also, the same attention tracking mechanism can be further optimized to simultaneously work with multiple subjects in a classroom using video footage from the CCTV cameras. Moreover, we have used human observed attention scores as ground truth-values as we currently do not have any dataset for measuring the attention span during online lectures. A standard dataset can help to evaluate the system's performance more reliably.